\documentclass[letterpaper, 10 pt, conference]{ieeeconf}  

\IEEEoverridecommandlockouts                              

\usepackage[compact]{titlesec}
\titlespacing{\section}{0.5pt}{0.5ex}{1ex}
\titlespacing{\subsection}{0pt}{0.1ex}{0ex}
\titlespacing{\subsubsection}{0pt}{0.1ex}{0ex}

\usepackage{graphics} 
\usepackage[demo]{graphicx}
\usepackage{amsmath} 
\usepackage{amssymb}  
\usepackage{caption}
\usepackage{subcaption}
\usepackage{array}
\usepackage{tabularx}
\usepackage{tensor}

\usepackage{comment}
\usepackage{xcolor}
\usepackage{cite}

\title{\LARGE \bf
Global Localization: Utilizing Relative Spatio-Temporal Geometric Constraints from Adjacent and Distant Cameras
}

\author{Mohammad Altillawi$^{1, 2}$, Zador Pataki$^{1, 3}$, Shile Li$^{1}$, and Ziyuan Liu$^{1}$
\thanks{$^{1}$AI Robotics \& Simulation, Huawei Munich Research Center, Germany.}
\thanks{$^{2}$Computer Vision Center, Universitat Aut\`onoma de Barcelona, Spain.}%
\thanks{$^{3}$Computer Vision and Geometry lab, ETH Zurich, Switzerland.}%
}

\begin{document}

\maketitle
\thispagestyle{empty}
\pagestyle{empty}

\begin{abstract}

Re-localizing a camera from a single image in a previously mapped area is vital for many computer vision applications in robotics and augmented/virtual reality.
In this work, we address the problem of estimating the 6 DoF camera pose relative to a global frame from a single image. We propose to leverage a novel network of relative spatial and temporal geometric constraints to guide the training of a Deep Network for localization. We employ simultaneously spatial and temporal relative pose constraints that are obtained not only from adjacent camera frames but also from camera frames that are distant in the spatio-temporal space of the scene. We show that our method, through these constraints, is capable of learning to localize when little or very sparse ground-truth 3D coordinates are available. In our experiments, this is less than 1\% of available ground-truth data. We evaluate our method on 3 common visual localization datasets and show that it outperforms other direct pose estimation methods.

\end{abstract}

\section{INTRODUCTION}

Estimating the position and orientation of a camera from a single image has been a central research topic in computer vision as it plays a crucial role in many tasks, such as robot navigation and Virtual Reality. With the advent of deep learning into computer vision, recent approaches have leveraged neural networks for data-driven pose estimation.
Direct pose regressors \cite{posenet, posenet+, poselstm, attloc, vidloc, mapnet, vlocnet, vipr} use pose labels to learn a direct mapping from image to pose. Although these methods regress pose in real-time, they were shown to be of limited localization accuracy compared to methods that use geometric information to localize \cite{limitations}.

\begin{figure}[ht]
\centering
\includegraphics[scale=0.5]{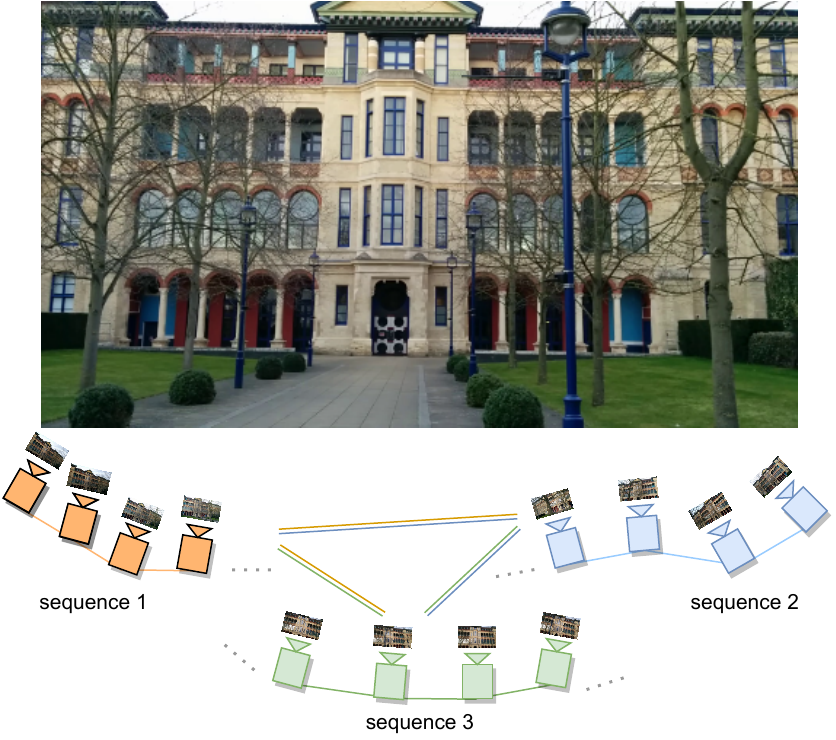}
  \caption{Our method utilizes relative geometric information from adjacent cameras as well as distant cameras to optimize the Deep Network weights for the benefit of global localization based on a single image. The proposed work applies relative pose constraints that are obtained from consecutive cameras along each sequence (denoted as single lines between two successive cameras) and from distant cameras across sequences that are far in the spatio-temporal space of the scene (denoted as double lines between two different sequences). These constraints are applied simultaneously for each training iteration.}
  \label{teaser}
\end{figure}

Most of the direct pose regressors learn from pose targets only. However, these methods ignore that, in most localization scenarios, additional geometric labels such as 3D coordinates of the target scene are available by default. Structure from motion (SfM) is the de-facto method 
 that is used to obtain ground-truth (GT) poses on outdoor scenes. SfM obtains camera poses and a 3D point cloud of the scene. In most indoor scenes where classic feature matching methods may fail because of repetitive structures or aliasing, depth sensors are used to track camera poses and reconstruct the scene. So, by default, obtaining GT poses for training direct pose regressors also delivers geometric scene information. If depth and poses are given, 3D scene coordinates can be obtained by back-projecting the depth to the 3D scene. Vice versa, if 3D scene coordinates are available, depth can be obtained by projecting the 3D coordinates into the camera frames relying on the available poses. As a result, 3D coordinates in a global reference frame, 3D coordinates in the camera frame (obtained from depth), and camera poses relative to the global reference frame are available for training. Direct pose regressors estimate the pose directly by regression without considering this geometric information. However, given the geometric scene information, a Deep Network can be trained to learn the geometry of the scene, which in return, can be used to localize. We exploit this geometric information in our method. Since the problem at hand is re-localizing a single image (only one image is available at inference time) in a previously mapped area, the geometric information of the scene is needed to obtain an absolute pose in metric units. These can be made available by either saving the 3D scene models of the environment (such as the output of SfM) or obtaining this information from the image while localizing. The latter is favorable, as the former requires a considerable additional memory, which is not anticipated when dealing with the problem of localizing a single image. Besides, utilizing saved GT 3D models at inference time imposes further computations, such as the ones that are driven by feature matching, which is needed to form correspondences between the query image and the 3D model.

In our work, we propose to train a Deep Network, which makes use of the readily available geometric labels, to localize a camera given a single image. At training time, the method employs the available camera poses in a global reference system, the 3D coordinates relative to this global reference system, and the 3D coordinates in the camera reference system to train a Deep Network to learn this geometric information. At inference time, the 3D coordinates in the global frame and those in the camera frame are estimated from a single image. We refer to these as map representations. To further constrain the map representations, we propose to utilize a network of simultaneous spatial and temporal relative geometric constraints. Specifically, we take advantage of multiple sequences of images wherein the sequences are taken at different times and in different scene spatial locations (Fig. \ref{teaser}). To fulfill this, we apply losses utilizing the relative poses not only between the successive samples of the same sequence (along the sequence) but also among different samples in different sequences (across sequences). Here, our method is not learning a temporal solution (localize the camera at a time instant given the camera poses/geometry at $N$ previous time instants). Instead, we use the sequences of images to impose more geometric constraints on the training process.

Given the two learned map representations, our method localizes the camera by aligning them using a classic rigid alignment method, Kabsch \cite{Kabsch}. However, localizing from the learned 3D map representations leads to inferior results due to imperfections in the learned maps. This is due to limited GT Labels availability or the presence of unseen objects in query images, such as dynamic objects or occlusions etc. To account for that, our method estimates weighting factors to guide the rigid alignment.

In summary, our contributions are:
\begin{itemize}

    \item We propose to utilize a network of relative geometric constraints along and across sequences for the benefit of global camera localization from a single image. These geometric constraints are computed as relative poses between camera frames that are adjacent or distant in space and time of the scene. Constraints from different sequences are applied simultaneously to update the Deep Network weights in each training iteration.

    \item Our method learns to localize when little or very sparse GT training 3D coordinates are available. We
     show that when less than 1\% of potential 3D coordinates are available.

    \item Our method outperforms state-of-the-art direct pose estimation methods.
    
\end{itemize}


\begin{figure*}[h]
\centering
\includegraphics[scale=0.7]{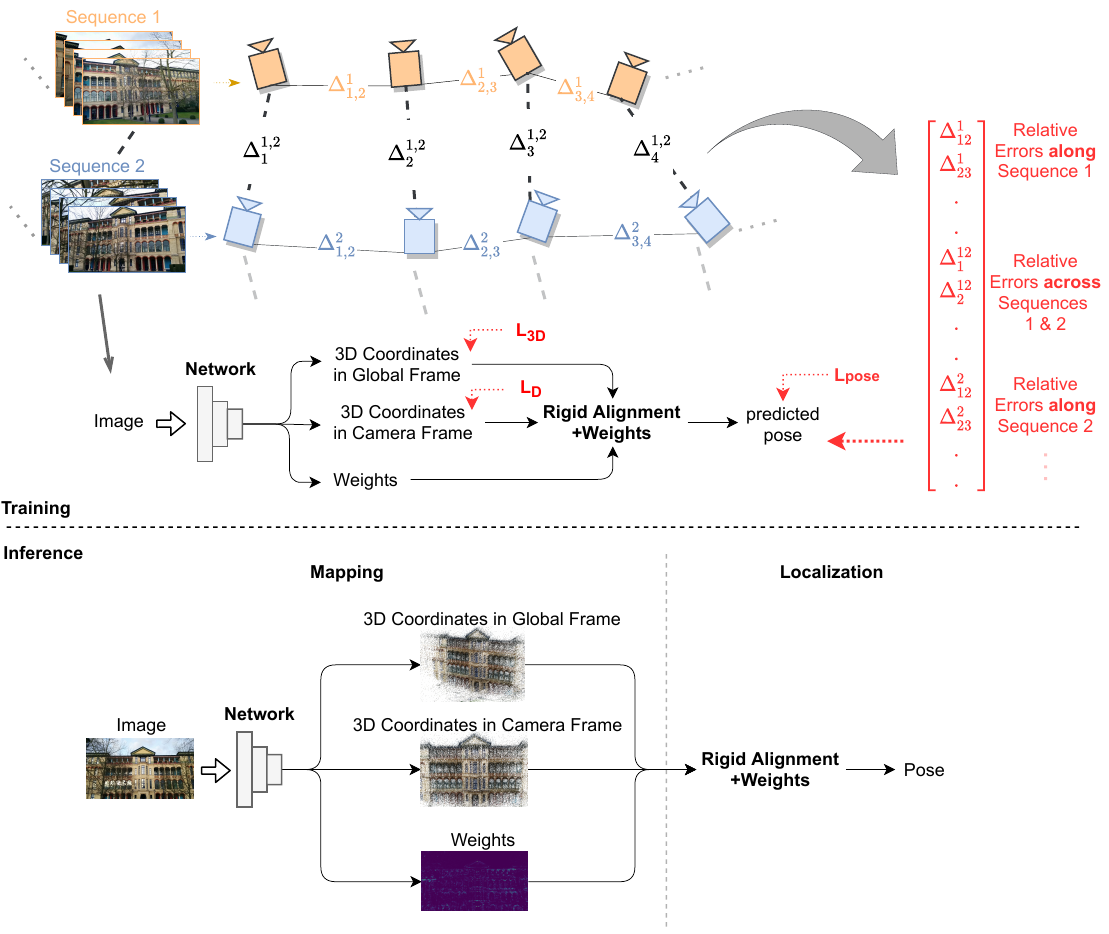}
  \caption{A diagram of our method. At training time, sequences of images are used to impose relative geometric constraints. For each image that is fed to the Deep Network, the following is predicted: the corresponding 3D coordinates in the global reference frame (Sec. \ref{globalcoords}), the depth which is used to estimate the 3D coordinates in the camera frame (Sec. \ref{cameracoords}), and weighting factors (Sec. \ref{heatmap}). Rigid alignment is used to align the two 3D point clouds to estimate a pose. $L_{3D}$, $L_{D}$, and $L_{pose}$ are losses that are applied during the training phase. In addition, relative pose errors along and across sequences are utilized as an additional signal to train the Deep Network (Sec. \ref{distant}). $\Delta^k_{i,j}$ stands for the relative pose error between two successive cameras $i$ and $j$ in the same sequence $k$ (along sequences). $\Delta^{k,l}_{i}$ stands for the relative pose error between the frame $i$ of each of the sequences of $k$ and $l$ (across sequences). At inference, an image is fed to the Deep Network to estimate the two map representations. Given that, a pose is obtained through weighted rigid alignment.}
  \label{main_diagram}
\end{figure*}


\section{RELATED WORKS}
Many recent works have addressed the camera global pose estimation. These works are categorized based on how the method computes a pose: direct or indirect.

\textbf{Indirect methods} are methods whose pose estimation consists of more than one step, like obtaining many pose hypotheses for further processing/refinement and/or condition on external dependencies like image retrieval. Compared to direct methods, they include additional probabilistic components such as database querying, hypotheses sampling, pose refinement, RANSAC \cite{ransac}, etc. Active Search \cite{activesearch}, a classic structure-based method, provides a prioritization scheme to establish matches between the extracted 2D features from a query image and the 3D points from a given 3D model and terminates correspondence search once enough matches are found. The pose is then obtained form the 2D-3D correspondences in a RANSAC framework \cite{ransac}. Instead of utilizing 3D model at inference, other methods \cite{dsac++} \cite{dsacstar} \cite{pixselect} learn the 3D scene and regress it at inference. Similar to classic methods, a pose is computed from the set of 2D-3D correspondences using perspective n-point pose solver in a RANSAC scheme \cite{ransac}. The pose is then refined with the inliers. PixLoc \cite{pixloc}, another indirect method, uses an image retrieval method to retrieve the most relevant database images to a query image, in addition to pose priors, and a reference 3D SfM model to estimate a pose. These methods obtain accurate poses and exceed direct methods.
In our work, the pose estimation is a single-step point cloud alignment.

\textbf{Direct methods} estimate camera pose in a deterministic way using a fixed pipeline without depending on external processes like retrieval from a database of images, matching features, sampling of pose hypotheses, and refinement of selected hypotheses using matched inlier correspondences. Pose regressors like PoseNet \cite{posenet} and followers falls into this category.

Direct methods estimate pose through regression \cite{posenet, posenet+, poselstm, attloc, vidloc, mapnet, vlocnet, vipr} or a single-step rigid alignment \cite{blanton2022structure}. Pose regression methods learn to map the input image directly into a 6 DoF output pose. A CNN Network learns a function that encodes the input into a latent feature. One or more regression layers map the latent image representation into the pose. These methods followed different learning strategies to encode beneficial information in the latent vector, such as LSTMs \cite{poselstm}, 3D model \cite{posenet+}, attention \cite{attloc}, graph neural networks \cite{gnn}, and sequence of images \cite{vidloc, mapnet, vlocnet, attloc, vipr, gnn, Seq_enhancement}. The common practice is that these methods model the geometric pose estimation problem as a regression task. In contrast to that, we estimate two geometric representations of the scene and use them to estimate the pose by aligning the two 3D representations in one step. Some of the methods that utilize sequence of images encode the temporal information along the same sequence consecutively through recurrent neural networks, which require the sequence of images also at inference \cite{vidloc, vipr, Seq_enhancement}. Opposite to that, we use information from sequences only for training and to optimize the two geometric map representations. At inference time, we localize from a single image. Furthermore, none of them utilizes simultaneous spatial and temporal geometric relative information along and across sequences. Another direct method that relies on rigid alignment for pose estimation is StructureAware \cite{blanton2022structure}. However, it learns the map from GT targets only, which may be of limited availability. In contrast to that, our method exploits close and distant spatio-temporal geometric information across the scene. These constraints are obtained from the available GT data without additional sensory inputs.


\section{METHOD}

\subsection{Overview}

This work addresses the problem of a single image pose estimation, i.e., estimating the pose of a camera with respect to a previously mapped area of a given global reference coordinate system from a single image. An overview of our method is shown in Figure \ref{main_diagram}.

During training, we use the target 3D global scene coordinates, the depth, and the GT poses to supervise the training of the Deep Network to estimate two representations of the map as seen by the image. The first is a set of 3D points with coordinates relative to the global coordinate frame of the scene and the second is a set of 3D coordinates as seen by the camera (relative to the camera frame). For the latter, the Deep Network estimates the depth, which is then back-projected given the camera intrinsic parameters. The two map representations are corresponding to each other, i.e., no matching is required to find 3D-3D correspondences. In addition, the Deep Network estimates a corresponding set of weighting factors. To estimate the camera pose, we use differentiable Kabsch algorithm \cite{Kabsch} to align the two maps according to the estimated weighting factors. The alignment is a single step closed-form solution. With a differentiable Kabsch algorithm, the whole Deep Network can be trained in an end-to-end manner
In the following, the components of the proposed method are described in details. For all the equations below, we denote the predicted counterpart of any GT term with a hat $\hat{}$.

\subsection{3D points in global coordinate frame} \label{globalcoords}

The proposed work estimates the 3D coordinates of the image pixels relative to the global reference system of the scene. This set of 3D coordinates forms the first map representation. To train the Deep Network to estimate a global map representation, the difference between the predicted and the available reference 3D scene coordinates is minimized. This is defined as follows:

\begin{equation}
L_{3D} = \frac{1}{M} \sum_{i}^{M} \parallel  \hat{\mathbf{x}}_i - \mathbf{x}_i \parallel_2,
\label{global_loss}
\end{equation}
where $X = { \{\mathbf{x}_i, ..., \mathbf{x}_M\}}$ is the set of the GT 3D global coordinates that corresponds to the set of predicted 3D coordinates $\hat{X} = { \{\hat{\mathbf{x}}_i, ..., \hat{\mathbf{x}}_M\}}$ and $M$ is the number of considered points for learning. We apply this loss where the 3D coordinates are available. For this purpose, we create a mask that indicates where the 3D coordinates are valid (available) and apply the loss where the mask is valid.
For the case of outdoor scenes, a small and very sparse 3D coordinates can be available. We show that our method is able to learn to localize when less than 1\% of GT 3D coordinates are available.

\subsection{3D points in camera coordinate frame} \label{cameracoords}

For learning the second map representation, i.e., the set of 3D coordinates from the perspective of the camera, the available depth labels are used to guide the Deep Network during training. For this goal, the Deep Network is supervised by a loss that minimizes the difference between the predicted and available GT depth as follows:

The depth is learned using $L_1$ loss:
\begin{equation}
L_{D} = \frac{1}{M} \sum_{i}^{M} |\hat{d}_i - d_i|,
\label{depth_loss}
\end{equation}
 where $\hat{d}_i$ and $d_i$ are the predicted and GT depth values, respectively. Each of them corresponds to pixel $i$ from the set $M$ of considered pixels.
Depth, if not directly available, can be obtained by projecting the 3D coordinates into the image using the GT poses. Since the depth or the 3D global coordinates can be obtained from each other, the mask of availability for both data is the same in most cases. 

 The depth is then transformed to 3D coordinates in the camera frame using $\mathbf{\hat{y}}_i = \hat{d}_i \mathrm{k^{-1}}u_i$, where $u_i$, $k$, $\hat{d}_i$, and $\mathbf{\hat{y}}_i$ denote the homogeneous pixel coordinates, the camera intrinsic matrix, the depth, and the corresponding point in the camera frame, respectively.

\subsection{Pose estimation} \label{rigid}
Given the two corresponding map representations, we use rigid alignment to align them. The two maps are explicitly corresponding to each other. We solve for the pose using Kabsch algorithm \cite{Kabsch}. In the training phase, we utilize its underlying differentiable singular value decomposition (SVD) to pass gradients from a pose loss to update the Deep Network weights.
To estimate the predicted translation vector $\hat{\mathbf{t}}$ and rotation matrix $\hat{\mathbf{R}}$, a cost function can be defined as:

\begin{equation}
\underset{\hat{\mathbf{R}},\hat{\mathbf{t}}} {\arg\min} \sum_{i}^{M}||\hat{\mathbf{x}}_i - \hat{\mathbf{R}}\hat{\mathbf{y}}_i - \hat{\mathbf{t}}||_2.
\label{non_weighted}
\end{equation}

$\hat{\mathbf{R}}$ and $\hat{\mathbf{t}}$ are recovered using a single step closed-form solution with singular value decomposition, which is parameter-free and differentiable.

Consequently, we define the pose loss as:

\begin{equation}
L_{pose} = L_{tr} + L_{rot},
\label{pose_loss}
\end{equation}
where $L_{tr}$ and $L_{tr}$ stand for the translation and the rotation losses, respectively. We define the translation loss as:

\begin{equation}
 L_{tr}(\mathbf{t}_{i}, \hat{\mathbf{t}}_{i}) =  \parallel  \mathbf{t} - \hat{\mathbf{t}} \parallel_2,
\label{translation}
\end{equation} 
where $\mathbf{t}$ and $\hat{\mathbf{t}}$ are the GT and predicted translations, respectively. We find that utilizing $L_2$ loss for the translation error supports the overall training of our method better than $L_1$. With different representations for the rotation, we found that the following definition: 
\begin{equation}
L_{rot}(\mathbf{R}_{i}, \hat{\mathbf{R}}_{i}) = \cos^{-1}(\frac{1}{2}(\mathrm{trace}(\hat{\mathbf{R}} \mathbf{R}^{-1}) -1))
    \label{rotation}
\end{equation}
works well in combination with the other losses defined in our work. $\mathbf{R}$ is the GT rotation matrix and $\hat{\mathbf{R}}$ stands for its predicted counterpart.

\subsection{Distant and adjacent spatio-temporal constraints:} \label{distant}

We introduce a novel network of simultaneous relative geometric constraints that cover the spatial and temporal distributions of the scene. The motivation is to constrain the Deep Network to better encode the two map representations for the best of pose estimation. For this goal, we apply geometric constraints not only along the consecutive camera frames but also across different sequences that are distant in space and time (of data collection). These sets of constraints are applied simultaneously. While the GT labels take care of the direct supervision of each frame, these relative geometric constraints introduce further training signals that cover the distant and adjacent spatio-temporal scene geometry.
We show in sections \ref{little} and \ref{effectiveness} the effectiveness of these constraints in learning to localize from little or sparse GT 3D coordinates labels.

For the mentioned purpose, we set these geometric constraints to relative poses between cameras that are consecutive (close in spatial location and time) and between cameras that are distant in spatial location and time (time of collection of the dataset). We are capable of using relative poses from distant cameras for training because we do not formulate the problem as temporal-based localization (i.e., localize frame $i$ from previous $N$ frames). With this setting, our work localizes from a single image at inference.

For a given image, its predicted pose is computed from aligning the two predicted map representations. The error between a predicted pose and a GT pose guides the training of the Deep Network by passing the gradients via the differentiable singular value decomposition module. A relative pose can be computed between any two cameras located at different locations in the scene. Consequently, we compute relative pose error as a difference between a relative pose computed from two predicted poses and its counterpart computed from the corresponding GT poses. By computing these relative poses between adjacent cameras in a sequence and, simultaneously, between cameras that are distant, we create a network of geometric constraints at every training iteration. These add additional training signals that further constrain the training.

For the equations below, we denote a pose with a rotation matrix $\mathbf{R}$ and a translation vector $\mathbf{t}$ as $\mathbf{T}$.
The relative pose between two camera frames $i$ and $j$ is defined as:

\begin{equation}
    ^{i}\mathbf{T}_{j} = \mathbf{T}_{i}^{-1}\cdot\mathbf{T}_{j},
\end{equation}
where $\mathbf{T}_{i}$ and $\mathbf{T}_{j}$ are the camera poses in the global frame for images $i$ and $j$, respectively. Consequently, the relative pose error for two camera frames $i$ and $j$ is defined as:

\begin{equation}
 L_{relpose}= L_{tr}(^{i}\mathbf{t}_{j}, ^{i}\hat{\mathbf{t}}_{j}) + \\ L_{rot}(^{i}\mathbf{R}_{j}, ^{i}\hat{\mathbf{R}}_{j}),
\label{relpose}
\end{equation}
where $^{i}\mathbf{t}_{j}$ and $^{i}\mathbf{R}_{j}$ are the translation and rotation parts of the GT relative pose $^{i}\mathbf{T}_{j}$, respectively. Similarly, $^{i}\hat{\mathbf{t}}_{j}$ and $^{i}\hat{\mathbf{R}}_{j}$ are, subsequently, the translation and rotation parts of the predicted relative pose $^{i}\hat{\mathbf{T}}_{j}$.

For K different sequences, each of N camera frames, the relative pose errors along the sequences are computed according to:
\begin{multline}
 L_{Along} = \sum_{k}^{K}\sum_{i}^{N-1}L_{tr}(^{i}_k\mathbf{t}_{i+1}, ^i_k\hat{\mathbf{t}}_{i+1}) + \\  \sum_{k}^{K}\sum_{i}^{N-1}L_{rot}(^{i}_k\mathbf{R}_{i+1}, ^{i}_k\hat{\mathbf{R}}_{i+1}),
 \label{LossAlongSeq}
  \end{multline}
where the term $^{i}_k\mathbf{t}_{i+1}$ means the translation vector of the relative GT pose between consecutive cameras $i$ and $i+1$ of sequence $k$. The same notation applies to the other terms in the equation.
Consequently, the relative pose errors across the sequences are computed according to:
\begin{multline}
 L_{Across} = \sum_{i}^{N}\sum_{k}^{K-1}L_{tr}(^{k}_i\mathbf{t}_{k+1}, ^k_i\hat{\mathbf{t}}_{k+1}) + \\ \sum_{i}^{N}\sum_{k}^{K-1}L_{rot}(^{k}_i\mathbf{R}_{k+1}, ^{k}_i\hat{\mathbf{R}}_{k+1}).
 \label{LossAcrossSeq}
 \end{multline}

Without confusing the terms of this equation with \eqref{LossAlongSeq} above, the term $^{k}_i\mathbf{t}_{k+1}$ denotes the translation vector of the relative GT pose between two distant cameras with index $i$, one in sequence $k$, and the other in sequence $k+1$. The same notation applies to the other terms in the equation.

\subsection{Weights} \label{heatmap}

To account for the elements that can degrade the localization performance, we predict a set of weights $W = { \{w_i, ..., w_M\}}$ to guide the rigid alignment. These elements can be unseen objects such as dynamic objects, occlusions, irrelevant scene areas such as the sky, or inaccuracies in the training data. These weights measure the contribution of each 3D-3D correspondence to the rigid alignment and are learned during training. For this purpose, we consider a weighted version of the rigid alignment.

In this section, we remove the distinction between predicted quantities (denoted previously by $\hat{.}$ ) and GT ones. The weighted minimization goal is defined as:

\begin{equation}
\underset{\mathbf{R},\mathbf{t}} {\arg\min} \sum_{i}^{M}w_i||\mathbf{x}_i - \mathbf{R}\mathbf{y}_i - \mathbf{t}||_2,
\label{weighted}
\end{equation}
where $w_i$ indicates how much a pair of points contribute to the pose estimation. The algorithm works as follows:
the translation $\mathbf{t}$ of the pose is removed by centering both point clouds:

\[ \boldsymbol{\mu}_{\mathbf{x}} =  \frac{\sum_{i} w_i\mathbf{x}_i}{\sum_{i} w_i},  \quad  \bar{\mathbf{X}} = \mathbf{X} - \boldsymbol{\mu}_{\mathbf{x}} \]

\[ \boldsymbol{\mu}_{\mathbf{y}} =  \frac{\sum_{i} w_i\mathbf{y}_i}{\sum_{i} w_i},  \quad  \bar{\mathbf{Y}} = \mathbf{Y} - \boldsymbol{\mu}_{\mathbf{y}}. \]

$\mathbf{R}$ and $\mathbf{t}$ are then recovered with SVD as follows:

\[\mathbf{U}\mathbf{S}\mathbf{V}^T = \mathrm{svd}(\bar{\mathbf{Y}}^{T}W\bar{\mathbf{X}}) \]
\[s = \mathrm{det}(\mathbf{VU}^T) \]
\[\mathbf{R} = \mathbf{V} \begin{pmatrix} 
	1 & 0 & 0 \\
	0 & 1 & 0\\
	0 & 0 & s \\
	\end{pmatrix} \mathbf{U}^T\]
\[\mathbf{t} = -\mathbf{R}\boldsymbol{\mu}_{\mathbf{y}} + \boldsymbol{\mu}_{\mathbf{x}}\]

Hence, to train the Deep Network to learn these weights, we fill in the predicted rotation and translation of \eqref{translation} and \eqref{rotation} by the values obtained from \eqref{weighted}.

\section{EXPERIMENT}

\begin{table*}[ht]
\caption{Results of the experiment of Sec. \ref{little}. The second row lists the average number of available GT 3D coordinates per image and the corresponding \% of the total number of possible GT points (6420) for output resolution of $60\times107$.}
\label{littleGT}
\begin{center}
\scriptsize
\begin{tabular}{c | c c c c c c c c}
Method & College & Hospital & Shop & Church\\
\hline
Available GT: count / \%  & 34 / 0.53\% & 15 / 0.24\% & 16 / 0.24\%  & 35 / 0.55\% \\
\hline
Baseline &	35.37m, 92.47°&	33.98m, 120.12°& 9.54m, 109.65°&	24.13m, 109.78°\\
DSAC* &	35.79m, 92.69°&	35.59m, 157.36°& 10.50m, 102.68°& 32.15m, 134.74°\\

\textbf{Ours}  & \textbf{0.52m, 1.39°} & \textbf{0.78m, 3.94°} & \textbf{0.42m, 1.71°} &	\textbf{0.94m, 4.55°} \\

\hline
\end{tabular}
\end{center}
\end{table*}

\subsection{Datasets}
We conduct our experiments on three common visual localization datasets: the outdoor Cambridge Landmarks \cite{posenet} and the indoor datasets: 7Scenes \cite{Shotton2013SceneCR} and 12Scenes \cite{12scenes}.

\textbf{Cambridge landmarks} is an outdoor relocalization dataset that contains RGB images of six large scenes, each covering a landmark of several hundred or thousand square meters in Cambridge, UK. The provided reference poses are reconstructed from SfM. The authors provide the train and test splits. We use the provided SfM models to obtain the GT 3D points for each image. Depth is obtained by projecting the 3D points into the camera frame using the GT poses. Following previous works \cite{poselstm,svspose,blanton2022structure, GposeNet}, we do not conduct experiments on the street and Court landmarks.

\textbf{7Scenes} is a RGB-D indoor relocalization dataset compromised of 7 scenes depicting difficult scenery such as motion blur, reflective surfaces, repeating structures, and texture-less surfaces. Several thousands frames with corresponding GT poses are provided for the train and test split of each scene. To compute the GT 3D global scene coordinates for training, we use the rendered depth maps from \cite{depth_7scenes} as these are registered to the RGB images.

\textbf{12Scenes} is an indoor dataset that consists of 12 RGB-D sequences. Compared to 7scenes, it covers larger indoor environments with smaller number of training images of several hundred frames for each scene. We obtain the 3D global coordinates in the same manner as for 7Scenes by using the rendered depth maps from \cite{depth_12scenes}.

\subsection{Architecture and Setup} \label{arch}
We implement the proposal using a single fully convolutional Deep Network with skip connections. The Deep Network takes a RGB input image. After three residual skip connections, the Deep Network branches out into three branches. Each branch corresponds to one of the predictions. The first is a 3 channels output that corresponds to the \textit{X, Y, and Z} coordinates in the global frame. The second is a one-channeled depth prediction. The third is also of one channel that store the weights for the rigid alignment. We apply stride-2 convolutions to downsample the input resolution by a factor of 8. We add Relu \cite{relu} as a non-linear activation after each layer except the last output layers. For the weights prediction, we use Sigmoid as activation for the last layer.

We resize the input images to a standard 480 px height and normalize them by mean and standard deviation. We do not apply any image scaling as augmentation. Thus, the input resolution is kept during the training. For each scene, the GT global 3D coordinates mean is subtracted from the predictions, which is then added at inference. During training, we apply, on the fly, color jittering and random in-plane rotations in the range [-30°, 30°].

We use the Adam optimizer with $\beta_1 = 0.9$, $\beta_2 = 0.999$, $\epsilon = 10^{-8}$, and a weight decay of $5\times10^{-4}$ and a learning rate of $10^{-4}$. All losses are applied with equal weighting factors of $1$.

For our experiments below, we obtain our network of distant spatio-temporal constraints from 2 sequences, each of 8 consecutive camera frames. The two sequences are sampled randomly for every training iteration. We reshape the input to be of dimension (B, C, H, W), where B, C, H, and W are the batch size, number of channels, height, and width of the image, respectively, with B being $16$ ($2\times8$).

Following other works, we report the pose error in the experiments below, as two components, median translation error (in meters) and median rotation error (in degrees).

\begingroup

\setlength{\tabcolsep}{2.6pt}

\begin{table*}[ht]
\caption{The table shows the results of the experiment of Sec. \ref{effectiveness}. Errors are reported as: median translation error (meters)/median rotation error (degrees). The second row reports, for each scene, the average percentage of available GT 3D coordinates of the total number of possible GT points per image. Best results are marked in bold.}
\label{effectOfDistantF}
\begin{center}
\scriptsize
\begin{tabular}{c | c c c c c c c c c c c c}
Method & Gates362 & Manolis & Gates381 & Lounge & Kitchen1 & Living1 & 5a & 5b & Bed & Kitchen2 & Luke & Living2\\
\hline
Available GT \%  & 91\% & 90\% & 92\%  & 95\%  & 90\%  & 95\%  & 94\%  & 96\%  & 95\%  & 93\%  & 93\%  & 91\%   \\
\hline
Baseline + PnP + RANSAC &	0.17/\textbf{2.12}& 0.10/4.13& 0.31/3.85& \textbf{0.06}/\textbf{2.07}& 0.11/\textbf{2.27}& \textbf{0.06}/1.81& 0.25/2.62& 0.10/2.51& 0.15/2.96& \textbf{0.04}/\textbf{1.87}& 0.24/4.16 & 0.07/2.77\\

Baseline + Rigid &	0.12/4.35& 0.23/9.43& 0.21/8.61& 0.14/4.83& 0.09/5.00& 0.11/3.46& 0.12/5.61& 0.17/6.75& 0.12/5.95& 0.08/3.78& 0.19/8.50 & 0.11/5.72\\

Ours + Rigid &	0.07/2.78& 0.10/4.47& 0.13/5.06& 0.11/3.53& 0.07/3.58& 0.08/2.60& 0.08/3.94& 0.13/5.39& 0.07/3.48& 0.06/2.63& 0.10/4.09& 0.09/4.04\\

Ours + Rigid + Weights & \textbf{0.05}/2.19& \textbf{0.07}/\textbf{2.77}& \textbf{0.08}/\textbf{3.26}& \textbf{0.06}/2.10& \textbf{0.05}/2.68& \textbf{0.06}/\textbf{1.75}& \textbf{0.06}/\textbf{2.23}& \textbf{0.07}/\textbf{2.24}& \textbf{0.04}/\textbf{1.88}& \textbf{0.04}/1.91& \textbf{0.08}/\textbf{2.85}& \textbf{0.06/2.30}\\

\hline
\end{tabular}
\end{center}
\end{table*}

\endgroup

\subsection{Learning from little or sparse GT} \label{little}
We show the effectiveness of our method for localization when little or very sparse GT data is available. 

We use the backbone described above (section \ref{arch}), which downsamples the input resolution by 8. For this resolution, we subsample the 3D GT coordinates by a factor of 8. We apply subsampling instead of interpolation in order to keep the accuracy of the GT. We perform the experiments on the outdoor Cambridge Landmarks, which provide sparse but enough GT 3D coordinates per image. Subsampling the sparse 3D coordinates by a factor of 8 results in a very small set of 3D coordinates. This results in a percentage of GT coordinates below 1$\%$ of the possible pixels that could have GT data for the Cambridge landmarks, 

We evaluate the localization performance of our method by comparing it to two methods. The first is the utilization of the GT data directly, which we take as a baseline. The second is DSAC* \cite{dsacstar}, a state-of-the-art visual localization method. To perform a fair comparison, we run the three methods on the same backbone using the same augmentation and optimizer settings (as described in section \ref{arch}).

\textbf{Baseline:} We apply \eqref{global_loss} to utilize the GT 3D coordinates, where they are available for training. For evaluation, we obtain poses using perspective-n-points algorithm \cite{p3p} in a RANSAC framework \cite{ransac} from all the 2D-3D correspondences (Open-CV implementation \cite{opencv_library}). The final pose is refined from the inliers. Since for this case, the weights and the 3D camera coordinates are not relevant, their corresponding branches are kept frozen (non-trainable).

\begin{table*}[ht]
\caption{Comparison against State-of-the-art localization methods on the 7scenes \cite{Shotton2013SceneCR}. Median translation (m) and rotation (°) errors (lower is better) with the improvements relative to the second best-reported quantity (underlined) are reported.}
\begin{center}
\scriptsize
\begin{tabular}{c c c c c c c c c c}
\hline
 & Methods & Chess & Fire & Heads & Office & Pumpkin & Kitchen & Stairs\\
\hline

Indirect & ActiveSearch \cite{activesearch} & 0.03, 0.87 & 0.02, 1.01 &  0.01, 0.82 &  0.04, 1.15 &  0.07, 1.69 &  0.05, 1.72 &  0.04, 1.01 \\
Indirect & DSAC* \cite{dsacstar} & 0.02, 1.10 & 0.02, 1.24 &  0.01, 1.82 &  0.03, 0.8 &  0.04, 1.34 &  0.04, 1.68 &  0.03, 1.16 \\
Indirect & PixLoc \cite{pixloc} & 0.02, 0.8 & 0.02, 0.73 &  0.01, 0.82 &  0.03, 0.82 &  0.04, 1.21 &  0.03, 1.20 &  0.05, 1.30 \\
\hline
&  & & & & & & &  & Average \\
\hline

Direct & Geo PoseNet \cite{posenet+} & 0.13, 4.48 & 0.27, 11.3 &  0.17, 13.0 &  0.19, 5.55 &  0.26, 4.75 &  0.23, 5.35 &  0.35, 12.4 & 0.23, 8.12\\
Direct & LSTM PoseNet \cite{poselstm} & 0.24, 5.77 &  0.34, 11.9 &  0.21, 13.7 &  0.30, 8.08 &  0.33, 7.00 &  0.37, 8.83 &  0.40, 13.7 & 0.31, 9.85
\\
Direct & Hourglass PN \cite{hourglass} & 0.15, 6.17 &  0.27, 10.8 &  0.19, 11.6 &  0.21, 8.48 &  0.25, 7.01 &  0.27, 10.2 &  0.29, 12.5 &  0.23, 9.54 \\
Direct & BranchNet \cite{branchnet} & 0.18, 5.17 &  0.34, 8.99 &  0.20, 14.2 &  0.30, 7.05 &  0.27, 5.10 &  0.33, 7.40 &  0.38, 10.3 & 0.29, 8.32 \\
Direct & GPoseNet \cite{GposeNet} & 0.20, 7.11 &  0.38, 12.3, &  0.21, 13.8, &  0.28, 8.83 &  0.37, 6.94 &  0.35, 8.15 &  0.37, 12.5 & 0.31, 9.95\\
Direct & MapNet \cite{mapnet} & 0.08, 3.25 &  0.27, 11.7 &  0.18, 13.3 &  0.17, 5.15 &  0.22, 4.02 &  0.23, 4.93 &  0.30, 12.1 & 0.21, 7.78
\\
Direct & AtLoc \cite{attloc} & 0.10, 4.07 &  0.25, 11.4 &  0.16, 11.8 &  0.17, 5.34 &  0.21, 4.37 &  0.23, 5.42 &  0.26, 10.5 & 0.20, 7.56\\
Direct & StructAware \cite{blanton2022structure} &	0.08, 2.17 & 0.21, 6.14 & 0.13, 7.93 & 0.11, 2.65 & 0.14, 3.34 & 0.12, 2.75 & 0.29, 6.88 & \underline{0.15}, \underline{4.55}\\
Direct & GNNPose \cite{gnn} & 0.08, 2.82 &  0.26, 8.94 &  0.17, 11.41 &  0.18, 5.08 &  0.15, 2.77 &  0.25, 4.48 &  0.23, 8.78 &  0.19, 6.32\\
Direct & AtLoc+ \cite{attloc} & 0.10, 3.18 &  0.26, 10.8 &  0.14, 11.4 &  0.17, 5.16 &  0.20, 3.94 &  0.16, 4.90 &  0.29, 10.2 &  0.19, 7.08\\
Direct & MapNet++ \cite{mapnet} & 0.10, 3.17 &  0.20, 9.04 &  0.13, 11.1 &  0.18, 5.38 &  0.19, 3.92 &  0.20, 5.01 &  0.30, 13.4 & 0.19, 7.29\\
Direct&  SeqEnhance \cite{Seq_enhancement} & 0.09, 3.28 &  0.26, 10.92 &  0.17, 12.7 &  0.18, 5.45 &  0.20, 3.66 &  0.23, 4.92 &  0.23, 11.3 & 0.19, 7.46 \\
Direct & \textbf{Ours}  &	\textbf{0.05, 1.52} & \textbf{0.15, 3.94} & \textbf{0.08, 6.24} & \textbf{0.11, 2.40} & \textbf{0.11, 2.52} & \textbf{0.11, 2.50} & \textbf{0.20, 5.56} & \textbf{0.12}, \textbf{3.53} \\
\hline
&  & & & & & & & Improvements: & \textbf{20\%, 22\%} \\

\hline
\end{tabular}
\end{center}
\label{indoor}
\end{table*}

\textbf{DSAC* \cite{dsacstar}:} We consider the RGB+3D model version of DSAC*, which requires a 3D model, camera poses, and images. DSAC* learns the 3D scene in two steps. In the first step, it utilizes the 3D GT coordinates to learn the geometry of the scene. In the second step, it implements a fully differentiable pose optimization. This applies a robust fitting of pose parameters using differentiable RANSAC \cite{dsacstar}. For the first stage, we take the results from the baseline above. In the second stage, the training from the baseline is further finetuned with the differentiable RANSAC of DSAC*. We use the DSAC* evaluation scripts that samples and refine the best-selected pose hypothesis for evaluation.

\textbf{Ours:} We train the Deep Network using our method by applying our relative pose constraints along \eqref{LossAlongSeq} and across \eqref{LossAcrossSeq} sequences simultaneously, the pose loss \eqref{pose_loss} in addition to utilizing the little available GT (\eqref{global_loss} and \eqref{depth_loss}). For evaluation, we obtain poses by aligning all the correspondences from the two map representations (3D coordinates in global and camera frames) using weighted rigid alignment.

We present the results in Tab. \ref{littleGT}. For the output resolution of 60$\times$107, which forms 6420 GT points; each training image frame hosts on average less than 1$\%$ of that. That is, 34 and 15 points out of 6420 per frame on average for college and hospital scenes, respectively.

\textbf{Analysis:}
Our geometric constraints from along and across sequences show a significant reduction of localization errors when learning from a very small number of 3D GT coordinates. This implies the effectiveness of our proposed work in learning to localize, provided small number of GT 3D coordinates. 
The baseline fails to learn map representations from this little GT data. Consequently, applying the fully differentiable pose optimization of DSAC* provides no more support for the training. Failure of DSAC* to adjust the training can be due to different reasons. DSAC* applies different augmentations, mainly scaling of the images, which requires scaling of the target GT data (implemented as an interpolation by DSAC*). In our experiments, we do not apply scaling. Additionally, DSAC* deploys additional training objectives (such as the reprojection errors) with multiple hyper-parameters to support the training. In our work, we do not apply reprojection error as a loss. Pairing DSAC* optimization functions as additional constraints for our work forms an interesting application to implement.

The baseline and DSAC* apply RANSAC to sample multiple pose hypotheses and refine the best one. In contrast, our pose estimation is a single direct step. We trained the baseline for a minimum of 600 epochs, while our work obtained the above results after 150 epochs. On GTX Titan X, our python implementation runs in 20.5ms at inference.

\begin{table*}[t]
\caption{Localization errors on the Cambridge Landmarks \cite{posenet}. Median translation (m) and rotation (°) errors (lower is better) with the improvements relative to the second best-reported quantity (underlined) are reported.}
\begin{center}
\scriptsize
\begin{tabular}{c c c c c c c c c}
\hline
 & Methods & College & Hospital & Shop & Church \\
\hline

Indirect & ActiveSearch \cite{activesearch} &	0.13, 0.22&	0.20, 0.36& 0.04, 0.21&	0.08, 0.25\\

Indirect & DSAC* \cite{dsacstar} &	0.15, 0.3&	0.21, 0.4& 0.05, 0.3&	0.13, 0.4\\

Indirect & PixLoc \cite{pixloc} &	0.14, 0.24&	0.16, 0.32& 0.05, 0.23&	0.10, 0.34\\

Indirect & PixSelect \cite{pixselect} &	0.14, 0.34&	0.14, 0.5& 0.06, 0.50&	0.09, 0.46\\
\hline
& & & & & & Average \\
\hline
Direct & PoseNet \cite{posenet} &	1.92, 5.40&	2.31, 5.38& 1.46, 8.08& 2.65, 8.08 & 2.08, 6.83 \\

Direct & Geo PoseNet \cite{posenet+} &	0.88, 1.04&	3.20, 3.29& 0.88, 3.78& 1.57, 3.32 & 1.63, \underline{2.86}\\

Direct & LSTM PoseNet \cite{poselstm} &	0.99, 3.65& 1.51, 4.29& 1.18, 7.44& 1.52, 6.68 & 1.30, 5.51\\

Direct & SVS-Pose \cite{svspose}  & 1.06, 2.81& 1.50, 4.03& 0.63, 5.73& 2.11, 8.11 & 1.32, 5.17\\

Direct & GPoseNet \cite{GposeNet}  &	1.61, 2.29& 2.62, 3.89& 1.14, 5.73& 2.93, 6.46 & 2.08, 4.59\\

Direct & MapNet \cite{mapnet} &	1.07, 1.89& 1.94, 3.91& 1.49, 4.22& 2.00, 4.53 & 1.62, 3.64\\

Direct & StructAware \cite{blanton2022structure}   &	1.19, 2.16& 1.11, \textbf{1.92}& 0.95, 6.82& 1.37, 4.45 & \underline{1.16}, 3.84\\

Direct & GNNPose \cite{gnn}  &	0.59, \textbf{0.65}& 1.88, 2.78& 0.50, 2.87& 1.90, \textbf{3.29} & 1.22, \textbf{2.40}\\

Direct & \textbf{Ours}  & \textbf{0.52}, 1.39 & \textbf{0.78}, 3.94 & \textbf{0.42, 1.71} &	\textbf{0.94}, 4.55 & \textbf{0.67}, 2,90\\
\hline
& & & & & Improvements: & \textbf{42\%}, - \\

\hline
\end{tabular}
\end{center}
\label{outdoor}
\end{table*}

\subsection{Effectiveness of the distant spatio-temporal constraints } \label{effectiveness}
This experiment shows the effectiveness of employing distant spatio-temporal constraints across sequences on pose estimation. We extend the baseline from the first experiment by including the branch that predicts the depth (consequently, the 3D camera coordinates). So that the baseline can learn the 3D scene, we apply this experiment on the indoor 12Scenes where GT 3D labels are available (dense). The dataset is challenging as low number of training images are available, but they cover relatively larger areas (compared to 7Scenes).

\textbf{Baseline + PnP + RANSAC:} we train the baseline to learn the two map representations (as our work) from the available dense GT 3D labels \eqref{global_loss} and \eqref{depth_loss}. For localization evaluation, we obtain poses by applying Perspective-n-Points algorithm \cite{p3p} in a RANSAC framework \cite{ransac} on all correspondences from the 2D image pixels and the 3D global coordinates. We use OpenCV implementation \cite{opencv_library} with 2000 iterations and a reprojection threshold of 10 pixels. The pose is then refined with the inlier correspondences.

\textbf{Baseline + Rigid:} we evaluate the baseline localization by computing the pose using rigid alignment of the two predicted 3D point clouds (3D coordinates in global frame and 3D coordinates in camera frame).

\textbf{Ours + Rigid:} We apply our distant spatio-temporal constraints (\eqref{LossAlongSeq} and \eqref{LossAcrossSeq}) and the Pose loss \eqref{pose_loss} on the baseline. We compute pose by aligning the two predicted 3D point clouds. In here, the correspondences are weighted equally.

\textbf{Ours + Rigid + weights:} We compute pose by weighted rigid alignment. Here, the correspondences are aligned according to the weights obtained from the weights branch.

\textbf{Analysis:}
Utilizing dense GT 3D coordinates helps the baseline to learn the maps representations and localization. By comparing Baseline+Rigid to Baseline+PnP+RANSAC we observe alternating performance between the two on the different scenes. On some scenes, PnP+RANSAC reports lower errors than rigid alignment, while on other scenes, the single step rigid alignment reports lower errors. Ours+Rigid (alignment with equal weightings) shows considerable improvements over Baseline+Rigid, showing the benefit of our added constraints. Ours+Rigid reduces the gap to Baseline+PnP+RANSAC on the scenes on which Baseline+PnP+RANSAC outperforms Baseline+Rigid and expands the gap on the other scenes, on which Baseline+Rigid outperforms Baseline+PnP+RANSAC. Considering the weights for the rigid alignment, we observe additional improvements on all the scenes compared to rigid alignment with equal weighting and improvements on majority of the scenes compared to Baseline+PnP+RANSAC.

\subsection{Comparison against state-of-the-art methods}
We compare our method against other global pose estimation methods. We list the localization results on the indoor 7Scenes dataset \cite{Shotton2013SceneCR} and the outdoor Cambridge Landmarks \cite{posenet} in tables \ref{indoor} and \ref{outdoor} respectively. We compare our results mainly against the direct methods since our method is direct. For completeness, we list the results of the indirect state-of-the-art methods. Looking at the indoor and outdoor results, we observe that our work reports the lowest localization errors on all indoor scenes and the lowest translation errors among the direct methods on the outdoor scenes.

Limitations: though our work learns the 3D scene as some of the indirect method, it reports relatively higher errors. This can be reverted to limitation in the ability of the rigid alignment to guide the learning of the weights. Indirect methods use a more robust outlier filtering approach.

\section{CONCLUSIONS}
This work addresses 6 DoF camera re-localization in a previously mapped area based on a single image. We propose to utilize relative geometric spatio-temporal constraints for training. These constraints are defined as relative poses and are obtained not only from adjacent and consecutive cameras but also from cameras that are distant in the spatio-temporal space of the scene. We apply these constraints simultaneously in every training iteration and we show their benefit in learning to localize from as little as below 1\% of 3D ground-truth coordinates. Our method outperforms other direct methods.
A potential direction for further works would be pairing our network of distant and adjacent geometric constraints with a differentiable RANSAC learning scheme and guided by a reprojection error objective function. 


\bibliographystyle{IEEEtran}
\bibliography{root}

\end{document}